\documentclass{article}

\PassOptionsToPackage{numbers, compress}{natbib}

\usepackage{arxiv}

\usepackage[utf8]{inputenc} 
\usepackage[T1]{fontenc}    
\usepackage[table]{xcolor}
\definecolor{citecolor}{HTML}{0071BC}
\definecolor{linkcolor}{HTML}{ED1C24}
\usepackage[colorlinks, citecolor=citecolor, linkcolor=linkcolor]{hyperref}
\usepackage{url}            
\usepackage{booktabs}       
\usepackage{amsfonts}       
\usepackage{nicefrac}       
\usepackage{microtype}      
\usepackage{lipsum}		
\usepackage{graphicx}
\usepackage{natbib}
\usepackage{doi}
\usepackage{wrapfig}
\usepackage{subcaption}
\usepackage{pifont}
\usepackage{multirow}
\usepackage{hyperref}
\usepackage{makecell}

\graphicspath{{resources/}}

\newcommand{\myparagraph}[1]{\noindent\textbf{#1}}

\newlength\savewidth\newcommand\shline{\noalign{\global\savewidth\arrayrulewidth
  \global\arrayrulewidth 1pt}\hline\noalign{\global\arrayrulewidth\savewidth}}
\newcommand{\tablestyle}[2]{\setlength{\tabcolsep}{#1}\renewcommand{\arraystretch}{#2}\centering\footnotesize}
\definecolor{baselinecolor}{gray}{.9}
\newcommand{\baseline}[1]{\cellcolor{baselinecolor}{#1}}
\newcolumntype{x}[1]{>{\centering\arraybackslash}p{#1pt}}
\newcolumntype{y}[1]{>{\raggedright\arraybackslash}p{#1pt}}
\newcolumntype{z}[1]{>{\raggedleft\arraybackslash}p{#1pt}}

\newcommand{\cmark}{\ding{51}}%
\newcommand{\xmark}{\ding{55}}%

\title{Predicting Brain Responses To Natural Movies With Multimodal LLMs}

\author{
Cesar Kadir Torrico Villanueva$^{1*}$ \quad
Jiaxin Cindy Tu$^{1,2}$\thanks{\quad Equal contributions, order of first four authors was randomized} \quad
Mihir Tripathy$^{1,3,*}$ \\
\textbf{Connor Lane$^{1,4*}$ \quad
Rishab Iyer$^{1,5}$ \quad
Paul S. Scotti$^{1,4}$}\\\\
$^{1}$Medical AI Research Center (MedARC) \quad
$^{2}$Psychological and Brain Sciences, Dartmouth College \\
$^{3}$Core for Advanced Magnetic Resonance Imaging (CAMRI), Baylor College of Medicine\\
$^{4}$Sophont \quad
$^{5}$Princeton Neuroscience Institute \quad 
}

\date{}

\renewcommand{\headeright}{Technical Report}
\renewcommand{\shorttitle}{\textit{MedARC} Team}

\begin{document}
\maketitle

\begin{abstract}
    We present MedARC's team solution to the Algonauts 2025 challenge. Our pipeline leveraged rich multimodal representations from various state-of-the-art pretrained models across video (V-JEPA2), speech (Whisper), text (Llama 3.2), vision-text (InternVL3), and vision-text-audio (Qwen2.5-Omni). These features extracted from the models were linearly projected to a latent space, temporally aligned to the fMRI time series, and finally mapped to cortical parcels through a lightweight encoder comprising a shared group head plus subject‑specific residual heads. We trained hundreds of model variants across hyperparameter settings, validated them on held‑out movies and assembled ensembles targeted to each parcel in each subject. Our final submission achieved a mean Pearson’s correlation of 0.2085 on the test split of withheld out-of-distribution movies, placing our team in fourth place for the competition. We further discuss a last-minute optimization that would have raised us to second place. Our results highlight how combining features from models trained in different modalities, using a simple architecture consisting of shared-subject and single-subject components, and conducting comprehensive model selection and ensembling improves generalization of encoding models to novel movie stimuli.
    Code is available at \href{https://github.com/MedARC-AI/algonauts2025}{\texttt{https://github.com/MedARC-AI/algonauts2025}}.
\end{abstract}

\section{Introduction}

The Algonauts Challenge is a biennial competition organized by the Algonauts Project in collaboration with the Conference on Cognitive Computational Neuroscience (CCN) aimed to advance our understanding of the human brain by encouraging researchers to develop the best encoding model for a given dataset. In computational neuroscience, encoding models refer to algorithms that take a stimulus as input and output predicted neural activations. Algonauts 2019–2023 focused on static images or short videos where models trained on deep convolutional networks were often adequate. However, humans experience the world through richly multimodal, temporally extended stimuli. The 2025 Algonauts challenge was designed to push the field beyond unimodal benchmarks by using long movies with synchronized video, audio and language. In collaboration with the CNeuroMod project, the organizers shared almost 80 hours of functional magnetic resonance imaging (fMRI) responses per subject.

Each of four subjects from the CNeuroMod project (sub-01, sub-02, sub-03, and sub-05) was scanned using fMRI while they watched 55 hours of Friends season 1 to 6 (\textit{friends s1-s6}) and 10 hours across four movies, including The Bourne Supremacy (\textit{bourne}), Hidden Figures (\textit{figures}), BBC documentary Life (\textit{life}), and The Wolf of Wall Street (\textit{wolf}). Their brain responses as time series were spatially normalized to the Montreal Neurological Institute (MNI) template \citep{brett_problem_2002}, and further downsampled to 1,000 functionally defined brain parcels spanning the left and right cerebrum \citep{schaefer2018local}. The test set was collected and processed with the same paradigm but with the four subjects watching 1) the friends season 7 (\textit{friends s7}) for the model-building phase and 2) six out-of-distribution (OOD) movie stimuli for the model-selection phase (for which fMRI responses were withheld). Each team was allowed up to ten submissions during the final model selection phase, which had a deadline of July 13, 2025.

\begin{figure}[t]
  \centering
  \includegraphics[width=4in]{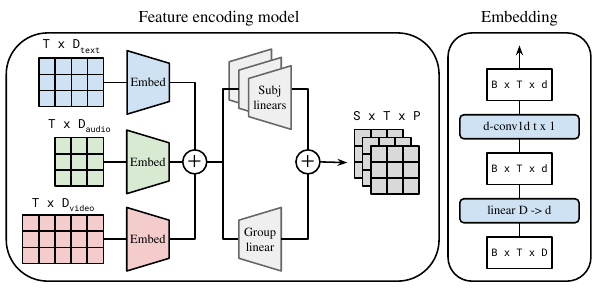}
  \caption{
    Encoding model architecture. The feature encoding model takes intermediate activations from a set of multimodal backbone models as inputs. These features are linearly projected and temporally aligned using independent ``embedding'' modules. The embeddings are then summed and passed through a linear fMRI prediction module consisting of a shared group head plus subject-specific heads. (\texttt{T} = number of time points, \texttt{D} = native feature dimension, \texttt{S} = number of subjects, \texttt{P} = number of parcels (1000), \texttt{d} = embedding dimension, \texttt{B} = batch size.)
  }
  \label{fig:arch}
\end{figure}

\section{Overview of our solution}

Overall, our three main contributions for success in the challenge consisted of (1) multimodal feature extraction using pre-trained models (Section~\ref{featureselection}), (2) parameter-efficient encoding architecture with temporal alignment with shared-subject and subject-specific adaptation (Section~\ref{sec:arch}), and (3) comprehensive model selection and ensembling to maximize generalization success to OOD movies (Section~\ref{ensembling}), including an ensemble model that was not submitted to competition in time which could have brought us to second place. Lastly, we discovered a strategy to use the returned Codabench scores for each movie to potentially increase the final scores in Section~\ref{optimizations}.

\section{Feature Selection and Extraction}
\label{featureselection}

To build a comprehensive predictive model of brain activity, we extracted representations from five distinct, deep neural network models. Each model was chosen to capture different facets of the audiovisual and linguistic content of the stimuli. The feature sets were extracted independently as high-dimensional feature vectors for each fMRI time point (known as repetition time or TR = 1.49 seconds). The general pipeline for each model involved segmenting the stimulus data (video, audio, and text if available) into chunks aligned with fMRI TR and processing them through the model to capture intermediate activations from the model's prespecified layer. This was determined empirically through a systematic search~\ref{sec:layer_selection_appd}.


\subsection{Multimodal Large Language Models (MLLMs)}
We utilized two state-of-the-art multimodal LLMs, InternVL3 \citep{zhu2025internvl3} and Qwen2.5-Omni \citep{xu2025qwen2} to capture integrated audiovisual and linguistic representation. The feature extraction for these models operates on a sliding window of the stimulus. For a given fMRI TR, a context window of the preceding 20 seconds of the video, audio (only for Qwen), and transcript data (if available) is processed in a single forward pass (the selection of this duration is justified in our ablation study in Appendix~\ref{sec:context_len_appd}.)

The intermediate activations from a pre-specified layer are captured for this entire sequence. The final feature vector for each TR is then derived by averaging the activation vectors over the token span corresponding to that specific TR's segment within the context window.



InternVL3 processes the 20-second context window as a sequence of image-text pairs. Specifically, the first frame of each TR, resized to 256 x 256 within the window, is selected and paired with its corresponding transcript chunk, if available. Qwen2.5-Omni samples the video segment at 2 frames per segment and resizes it to 256×384. The audio is sampled at 16 kHz audio waveform and fused with the available transcript to create a unified input representation.

\subsection{Specialized Unimodal Models}
To capture modality-specific information with high fidelity, we included features from specialized audio, language, and vision models. The general procedure for these models was to process their respective modality (video, audio, or text) to produce a single feature vector for every fMRI TR. This was achieved by passing the prepared input through the model, capturing activations from a target layer, and then applying a model-specific averaging or pooling strategy to produce a fixed-size vector per TR.

\textbf{V-JEPA2} (Vision-based Joint Embedding Predictive Architecture) \citep{assran2025v} is a non-generative, self-supervised vision model that learns an abstract understanding of world dynamics by predicting representations of masked-out video parts in a latent space. For each TR, its corresponding video segment was processed by selecting 15 frames, which were passed through the model's encoder. The final feature vector was produced by averaging the resulting image patch embeddings.

\textbf{Whisper} (Large-v3) \citep{radford2023robust} is an encoder-decoder transformer whose encoder learns robust representations of speech and general acoustics from large-scale pre-training for speech recognition. For each TR, its 16 kHz mono audio segment was passed through the Whisper encoder. The final feature vector was computed by averaging the temporal sequence of the resulting output activations.

\textbf{Llama 3.2} (3B) \citep{grattafiori2024llama} is a lightweight multilingual text-only language model. For this model, we processed the entire transcript for a stimulus in a single forward pass. The token spans corresponding to each TR were identified using character offsets, and the final feature vector for each TR was computed by averaging the activations of all tokens within its identified span.

\section{Model Architecture}
\label{sec:arch}

Our feature encoding model (Figure~\ref{fig:arch}) is made up of a feature embedding stage and an fMRI prediction stage. In the feature embedding stage, the extracted features from each input backbone model are first linearly projected frame-by-frame from their native dimension to a small latent dimension. The projected feature time series are then temporally aligned using depth-wise 1D convolutions \cite{howard2017mobilenets,liu2022convnet}, learned separately for each input feature. Finally, the embeddings from each backbone are summed together.
The fMRI prediction stage consists of a group linear prediction head that is shared across the four subjects, as well as subject-specific ``residual'' heads. The group and subject heads are applied to the embedding time series frame-by-frame, and their results are added together.

\myparagraph{Model training details.} In this section, we focus on a representative ``default'' model to understand baseline model performance. In Section~\ref{ensembling}, we develop an ensemble of different model variants to achieve our final submission score.
Our default model training setup uses all five backbone features described in Section~\ref{featureselection}. We used a default embedding dimension of 192 and convolution kernel width of 45 TRs (67 seconds). The total number of trainable parameters was 3.5M. Our default training data mixture included \textit{friends s1-5}, \textit{bourne}, and \textit{wolf}. We held out \textit{friends s6}, \textit{figures}, and \textit{life} for testing. We used \textit{figures} as a validation set for early stopping. We trained our models using AdamW for a maximum of 1200 steps with a batch size of 16 and a temporal sequence length of 64 TRs. Wall clock training time was under 2 minutes using a single NVIDIA H100 GPU ($<$1 GB memory usage).
 


\subsection{Default baseline model performance}

\begin{wrapfigure}{r}{0.4\textwidth}
  \centering
  \includegraphics[width=0.4\textwidth]{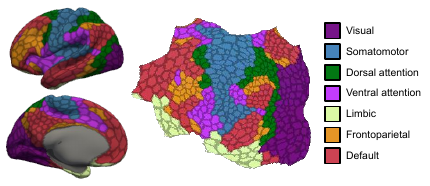}
  \caption{Schaefer 1000 parcellation with Yeo networks \citep{schaefer2018local} shown on the HCP inflated surface \citep{glasser2013minimal} and pycortex flat map \citep{gao2015pycortex}.}
  \label{fig:yeo}
\end{wrapfigure}
\begin{figure}[t]
  \centering
  \includegraphics[width=1.0\textwidth]{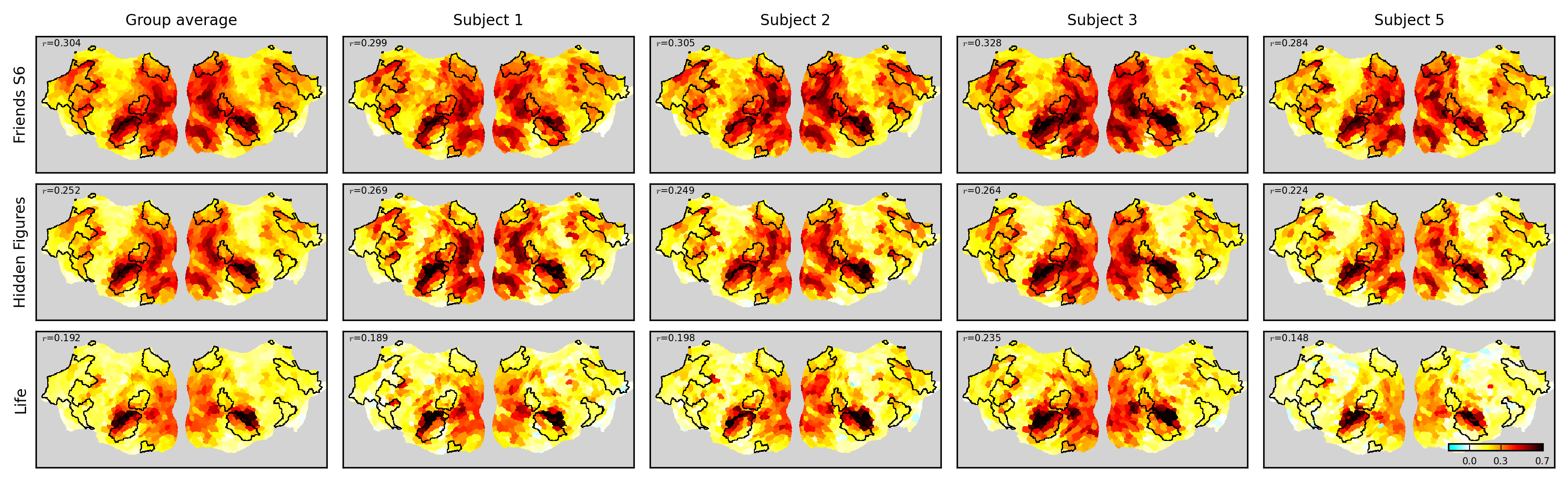}
  \caption{
    Feature encoding performance of our default model. Values are Pearson $r$ encoding accuracy maps computed on held-out validation movies. Black outline indicates the boundary of the Yeo "default" resting-state network \citep{schaefer2018local}.
  }
  \label{fig:maps}
\end{figure}

Figure~\ref{fig:maps} shows the performance of our default baseline model, without ensembling, which achieves a final OOD test score of 0.203. We observe robust prediction performance in the visual cortex as well as lateral prefrontal and temporal brain areas. (See Figure~\ref{fig:yeo} for reference.) Consistent with classic work mapping fMRI responses to naturalistic stimuli, e.g.~\citet{hasson2004intersubject,huth2012continuous}, prediction performance is especially strong in temporal areas responsive to sound, speech, and language, as well as inferotemporal and lateral occipital areas responsive to people, places, and action. We observe reliable albeit weaker prediction in parts of the dorsal and frontoparietal attention networks. Interestingly, prediction performance is notably weaker in the movie \textit{life} compared to \textit{friends s6} and \textit{figures}, especially in visual and frontoparietal areas. Unlike the other movies, \textit{life} is a narrated nature documentary without human characters or storylines.

\subsection{Ablation experiments}
\label{sec:ablations}

We evaluate the components of our default model in a set of ablation experiments summarized in Table~\ref{tab:ablations}.

\myparagraph{Convolution kernel size.} The fMRI signal has a temporally delayed and blurred response compared to the underlying neural signal and movie stimulus. Consistent with this, we find that temporally aligning the features with the fMRI is crucial for good performance (Table~\ref{tab:kernel_size}). We find that a large kernel width of 45 TRs (67 seconds) is best. Interestingly, this is much longer than the typical hemodynamic response. This suggests that aggregating features over a long context window is beneficial for predicting fMRI.

\myparagraph{Convolution kernel type.} We explore a range of convolution kernel types in Table~\ref{tab:kernel_type}. All kernel types use depthwise 1D convolution. The ``causal'' kernel uses a causal convolution mask so that predictions depend only on earlier (in time) features. The ``positive'' kernel constrains weights to be non-negative by applying absolute value to the weights. The ``tied'' kernel shares the same single-channel filter across all input embedding channels. We find that all kernel variants achieve accurate prediction, although the default standard depthwise convolution performs marginally better.

\myparagraph{Embedding dimension.} We find that an embedding dimension of 192 performs best (Table~\ref{tab:embed_dim}). However, our model achieves strong performance even with much smaller embedding dimensions. This suggests that although the fMRI data are high-dimensional, most of the predictable stimulus-driven activity lies near a low-dimensional (linear) manifold.

\myparagraph{Multi-subject training.} A key aspect of our approach is to leverage the aligned fMRI responses of multiple subjects to the same stimuli in order to jointly train a multi-subject encoding model. In Table~\ref{tab:multi_sub}, we find that this provides a $\Delta r= 0.01$ improvement over single subject training, with slightly more improvement for the OOD movie \textit{life}.

\myparagraph{fMRI prediction head architecture.} Another aspect of our model architecture is the fMRI prediction head, which we factorized into a shared group head plus subject-specific residual heads. We find that the subject-specific residual heads provide a clear improvement over just the group head (Table~\ref{tab:pred_head}). Adding the group head does not provide any meaningful benefit on top of just the subject-specific heads. The group head alone still predicts well, suggesting that much of the predictable signal is shared across people.

\myparagraph{Input feature set.} In Table~\ref{tab:features}, we test the impact of removing various features from the input feature set. In the first block of rows, we look at the performance of each backbone model feature individually. We observe that as expected, the multimodal modals InternVL3 (image, text) and Qwen2.5 (image, text, audio) achieve the best individual performance. Qwen2.5 performs especially well on the OOD movie \textit{life}. V-JEPA2 and Llama 3.2 both achieve strong individual performance, while Whisper performs notably worse on its own.
In the second block of rows, we look at the leave-one-model-out performance. We find that each backbone feature contributes to the overall encoding model performance (i.e.\ performance drops when each model is left out.) Llama 3.2 seems to have a strong, unique contribution, while Whisper has the weakest contribution.

\begin{table*}[t]
\centering
\subfloat[
\textbf{Kernel size}. Temporally aligning features with 1D convolution is key.
\label{tab:kernel_size}
]{
\begin{minipage}{0.29\linewidth}{\begin{center}
\tablestyle{4pt}{1.05}
\begin{tabular}{x{24}x{24}x{24}x{24}}
width & s6 & figures & life \\
\shline
none & 0.200 & 0.146 & 0.098 \\
9 & 0.282 & 0.230 & 0.177 \\
17 & 0.291 & 0.240 & 0.183 \\
45 & \baseline{\textbf{0.304}} & \baseline{\textbf{0.252}} & \baseline{\textbf{0.192}} \\
65 & 0.303 & 0.251 & \textbf{0.192} \\
\end{tabular}
\end{center}}\end{minipage}
}
\hspace{2em}
\subfloat[
\textbf{Kernel type}. Similar performance for different conv1d kernel variants.
\label{tab:kernel_type}
]{
\begin{minipage}{0.29\linewidth}{\begin{center}
\tablestyle{4pt}{1.05}
\begin{tabular}{y{28}x{24}x{24}x{24}}
case & s6 & figures & life \\
\shline
causal & 0.301 & 0.244 & 0.192 \\
positive & 0.299 & 0.246 & 0.187 \\
tied & 0.302 & 0.251 & \textbf{0.194} \\
default & \baseline{\textbf{0.304}} & \baseline{\textbf{0.252}} & \baseline{0.192} \\
\multicolumn{4}{c}{~}\\
\end{tabular}
\end{center}}\end{minipage}
}
\hspace{2em}
\subfloat[
\textbf{Embed dimension}. Strong performance even for small embed dimension.
\label{tab:embed_dim}
]{
\begin{minipage}{0.29\linewidth}{\begin{center}
\tablestyle{4pt}{1.05}
\begin{tabular}{x{24}x{24}x{24}x{24}}
dim & s6 & figures & life \\
\shline
32 & 0.298 & 0.249 & 0.190 \\
64 & 0.298 & 0.252 & 0.191 \\
128 & 0.303 & \textbf{0.253} & 0.194 \\
192 & \baseline{\textbf{0.304}} & \baseline{0.252} & \baseline{0.192} \\
256 & 0.300 & 0.252 & \textbf{0.195} \\
\end{tabular}
\end{center}}\end{minipage}
}
\\
\centering
\vspace{.3em}
\subfloat[
\textbf{Multi subject}. Jointly training on multiple subjects improves performance.
\label{tab:multi_sub}
]{
\centering
\begin{minipage}{0.29\linewidth}{\begin{center}
\tablestyle{4pt}{1.05}
\begin{tabular}{y{40}x{24}x{24}x{24}}
case & s6 & figures & life \\
\shline
single sub & 0.294 & 0.243 & 0.179 \\
multi sub & \baseline{\textbf{0.304}} & \baseline{\textbf{0.252}} & \baseline{\textbf{0.192}} \\
\multicolumn{4}{c}{~}\\
\end{tabular}
\end{center}}\end{minipage}
}
\hspace{2em}
\subfloat[
\textbf{Prediction head}. Subject-specific heads are necessary (and sufficient).
\label{tab:pred_head}
]{
\begin{minipage}{0.29\linewidth}{\begin{center}
\tablestyle{4pt}{1.05}
\begin{tabular}{y{46}x{24}x{24}x{24}}
case & s6 & figures & life \\
\shline
group only & 0.273 & 0.232 & 0.183 \\
sub only & 0.302 & \textbf{0.253} & \textbf{0.194} \\
sub $+$ group & \baseline{\textbf{0.304}} & \baseline{0.252} & \baseline{0.192} \\
\end{tabular}
\end{center}}\end{minipage}
}
\\
\centering
\vspace{.3em}
\subfloat[
\textbf{Feature set}. First block shows each feature's individual performance. Second block shows leave one feature out performance.
\label{tab:features}
]{
\begin{minipage}{0.6\linewidth}{\begin{center}
\tablestyle{4pt}{1.05}
\begin{tabular}{x{24}x{24}x{24}x{24}x{24}x{24}x{24}x{24}}
internvl & qwen & vjepa & whisper & llama & s6 & figures & life \\
\shline
\cmark & \xmark & \xmark & \xmark & \xmark & 0.245 & 0.188 & 0.116 \\
\xmark & \cmark & \xmark & \xmark & \xmark & \textbf{0.276} & \textbf{0.228} & \textbf{0.168} \\
\xmark & \xmark & \cmark & \xmark & \xmark & 0.229 & 0.177 & 0.101 \\
\xmark & \xmark & \xmark & \cmark & \xmark & 0.173 & 0.106 & 0.091 \\
\xmark & \xmark & \xmark & \xmark & \cmark & 0.230 & 0.177 & 0.134 \\
\hline
\xmark & \cmark & \cmark & \cmark & \cmark & 0.300 & 0.251 & 0.188 \\
\cmark & \xmark & \cmark & \cmark & \cmark & 0.301 & 0.250 & 0.188 \\
\cmark & \cmark & \xmark & \cmark & \cmark & 0.298 & 0.247 & \textbf{0.193} \\
\cmark & \cmark & \cmark & \xmark & \cmark & \textbf{0.303} & \textbf{0.253} & 0.192 \\
\cmark & \cmark & \cmark & \cmark & \xmark & 0.290 & 0.243 & 0.181 \\
\hline
\cmark & \cmark & \cmark & \cmark & \cmark & \baseline{\textbf{0.304}} & \baseline{\textbf{0.252}} & \baseline{\textbf{0.192}} \\
\end{tabular}
\end{center}}\end{minipage}
}
\caption{\textbf{Ablation experiments} training on \textit{friends s1-5}, \textit{bourne}, \textit{wolf}, and testing on \textit{friends s6}, \textit{figures}, \textit{life}. \textit{figures} was used as the validation set for early stopping. Default settings are marked in \colorbox{baselinecolor}{gray}.
}
\label{tab:ablations} \vspace{-.5em}
\end{table*}

\section{Model Selection and Parcel-wise Ensemble Strategy}
\label{ensembling}

For competition purposes, we employed a straightforward model selection and parcel-wise ensemble strategy to enhance performance~\cite{caruana2004ensembleselection}. Initially, we defined a comprehensive hyperparameter search space, varying parameters such as learning rate, weight decay, encoder kernel size, batch size, and embedding dimensions, among others. Additionally, multiple feature sets were prepared, including both individual model features and combinations thereof.

\begin{figure}[]
  \centering
  \includegraphics[width=0.78\textwidth]{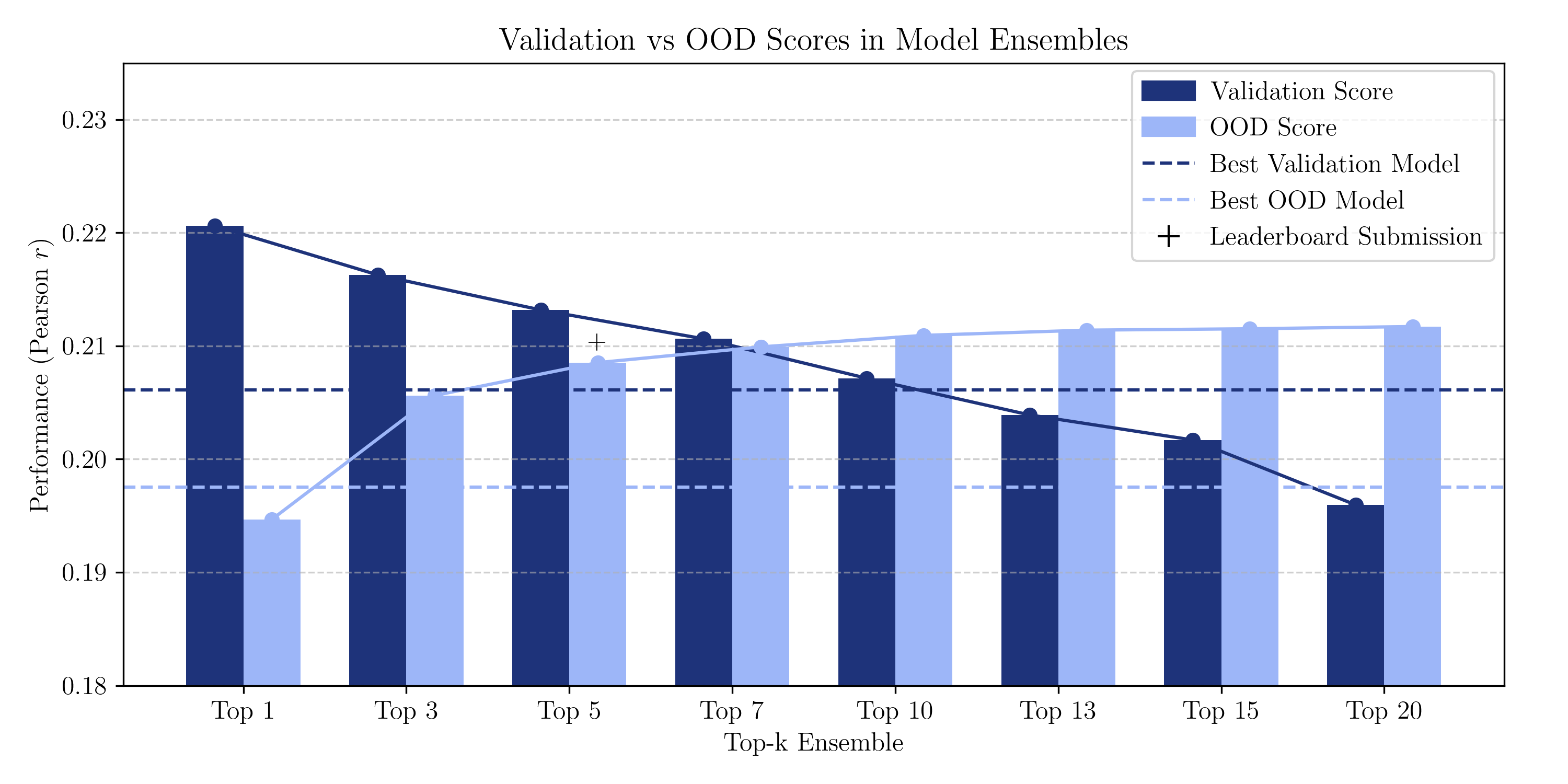}
  \caption{
    Comparison of ensemble performance across validation and out-of-distribution (OOD) data.\protect\footnotemark A total of 49 models were validated on \textit{life} and \textit{bourne}. Bars indicate Top-$k$ ensemble scores. Dashed lines represent the performance of the best single model, while the `+' symbol denotes the best score submitted to the leaderboard.
  }
  \label{fig:ensemble_scores}
\end{figure}

Subsequently, we randomly sampled multiple configurations from this hyperparameter space and trained each feature set using these configurations, reserving specific data (\textit{life} and \textit{bourne}) as the validation dataset (see Appendix~\ref{appendix:validation_set} for an ablation on validation set choice). For each parcel in each subject, we selected the top-$k$ models based on their performance on the validation dataset. We then averaged the fMRI activity predictions from these top-$k$ models for the OOD movies to produce the final prediction for submission, a procedure shown theoretically and empirically to reduce variance and improve generalization \cite{breiman1996bagging}. More complex variants of this strategy (e.g., random ROI-ensembles, weighted averaging) have been successfully applied to fMRI decoding tasks~\cite{yoshimoto2024ensembletraits,khosla2019ensemblecnn} and previous Algonauts editions~\cite{yang_memory_2023}. The results are summarized in Figure \ref{fig:ensemble_scores}.


The ensemble involved 49 models and provided a substantial increase in OOD performance of 0.011 for top-$k$=5, compared to the best single model, achieving a final leaderboard submission score of 0.208529. 

\subsection{Scaling Ensembles Further Improves Generalization}

While our official submission employed a Top-$k=5$ ensemble, increasing the ensemble size led to a consistent improvement in OOD generalization. In particular, a Top-$k=20$ ensemble achieved a Pearson correlation of 0.211717 on the OOD set. This result, if submitted, would have placed our entry in second place overall.

Unfortunately, this optimization was not available in time for the official competition. Nevertheless, it highlights the value of large-scale ensembling when sufficient validation and model diversity are available. This finding aligns with classical ensemble learning theory, where increasing ensemble size often leads to reduced variance and enhanced generalization, provided that individual models contribute diverse errors~\cite{breiman1996bagging}.

\footnotetext{Due to the 10-submission limit during the official competition, only two of the OOD results shown in this plot were actually submitted. The remaining results were retrospectively computed during the post-challenge phase.}

\section{Final thoughts: using Codabench scores to select the best model for each movie}
\label{optimizations}
The competition performance score on the leaderboard was calculated based on the average performance across six out-of-distribution movies that had some distinct characteristics from the training movies. For example, the movie "Chaplin" has no dialogue and is in black and white, in contrast to all training data. This posed a challenge to use the trained model with our previously identified optimal feature sets that include a language model. We first started by using dummy features for the Llama language model (using dummy transcripts where each line is just the time stamp). Using all "friends" and "movie10" movies as training data, our model's performance on "Chaplin" was 0.2205, 0.1467, 0.2057, and 0.1283 for the subjects 1, 2, 3, and 5, respectively. On a separate submission, we replaced the performance on Chaplin and trained with all "friends" and "movie10" movies without "Life" using a different model trained without Llama features, and saw the performance changed to 0.2394, 0.1412, 0.2004, and 0.1409, with a marked increase for subjects 1 and 5. In retrospect, the average performance score on "Chaplin" across subjects for the model without Llama features is higher than that of the top 3 submissions on the leaderboard.

Similarly, we noticed that the submissions from the ensemble solutions produced better scores for some movies, but not for others. Since the out-of-distribution data was so diverse, certain feature sets/training parameters may generalize better to specific out-of-distribution movies. Since Codabench provides the score for each movie, we were able to download the performance scores for each movie of each subject and select the best-performing model prediction among all our past submissions. This combined prediction was also not submitted in time for the competition, which could have resulted in a further increase in performance scores (average performance of r = 0.2105 across four subjects, 10\% increase from our best submission performance r = 0.2085). 

This optimization was only possible because Codabench provided movie-specific scores on the test set. While this approach was permitted under the competition rules, it partially circumvented the organizers’ goal of a single generalizable encoding model. A future direction could explore whether systematic biases exist—e.g., if certain feature sets consistently generalize better to specific movie types—and whether feature optimization could be automated based on intrinsic properties of the test movie without access to the test scores.

\section{Conclusion}
This technical report summarizes the modeling approach and experimentation that went into MedARC's fourth-place submission to the Algonauts 2025 challenge. We use multimodal feature extraction, a shared-subject architecture, and ensemble strategy to optimize fMRI encoding model generalization to out-of-distribution stimuli. We describe our model and feature set ablations which may inform the development of future brain encoding models. This work contributes to our growing understanding of how the brain represents everyday, multimodal human experiences.

\bibliographystyle{unsrtnat}
\bibliography{references}

\clearpage
\appendix

\section{Layer selection for feature encoding}
\label{sec:layer_selection_appd}
This ablation study details the procedure for determining the optimal layer from each of the five models for our final encoding model. We first conducted a broad feature extraction, generating stimuli feature sets from multiple candidate layers spanning the architectural depth of each model. Each feature set was then used to independently train an encoding model on our primary training corpus (\textit{Friends} S1-5, \textit{wolf}, and \textit{bourne}). The performance of each layer-specific model was subsequently evaluated on a held-out in-distribution dataset (\textit{friends S6}) and two out-of-distribution (OOD) datasets (\textit{life} and \textit{figures}). The final layer reported for each model was the one whose features yielded the highest predictive performance in this ablation, ensuring our final feature space was composed of the most neurally-relevant representations from our extracted pool.

\begin{table}[h!]
\centering
\caption{Layer-wise Pearson Correlation Scores for all models}
\label{tab:all_model_performance}
\begin{tabular}{lcccc}
\toprule
\textbf{Model} & \textbf{Layer} & \multicolumn{3}{c}{\textbf{Pearson Correlation Score}} \\
\cmidrule(lr){3-5}
& & \textit{Friends S6} & \textit{Figures} & \textit{Life} \\
\midrule
\multicolumn{5}{c}{\textit{Multimodal Large Language Models}} \\
\midrule
\multirow{3}{*}{InternVL3-8B} & 10 & 0.2627 & 0.2061 & 0.1529 \\
& 15 & 0.2739 & 0.2197 & 0.1627 \\
& \textbf{20} & \textbf{0.2753} & \textbf{0.2219} & \textbf{0.1685} \\
\midrule
\multirow{4}{*}{InternVL3-14B} & 20 & 0.2394 & 0.1784 & 0.0960 \\
& \textbf{30} &\textbf{0.2431} & \textbf{0.1837} & \textbf{0.1102} \\
& 40 & 0.2338 & 0.1822 & 0.1011 \\
& 47 & 0.2300 & 0.1764 & 0.0950 \\
\midrule
\multirow{3}{*}{Qwen2.5-Omni 3B} & 10 & 0.2641 & 0.2117 & 0.1489 \\
& 15 & 0.2712 & 0.2185 & 0.1511 \\
& \textbf{20} & \textbf{0.2746} & \textbf{0.2239} & \textbf{0.1671} \\
\midrule
\multirow{4}{*}{Qwen2.5-Omni 7B} & 10 & 0.2681 & 0.2095 & 0.1483 \\
& 15 & 0.2702 & 0.2181 & 0.1607 \\
& \textbf{20} & \textbf{0.2703} & \textbf{0.2189} & \textbf{0.1609} \\
& 25 & 0.2584 & 0.2025 & 0.1425 \\
\midrule
\multicolumn{5}{c}{\textit{Specialized Unimodal Models}} \\
\midrule
\multirow{5}{*}{VJEPA2 ViT-G} & 5  & 0.1721 & 0.1119 & 0.0449 \\
& 15  & 0.1910 & 0.1275 & 0.0562 \\
& 25  & 0.2256 & 0.1724 & 0.0964 \\
& 35 & 0.\textbf{2333} & 0.1834 & 0.1089 \\
& \textbf{layernorm} & 0.2266 & 0.\textbf{1839} & 0.\textbf{1177} \\
\midrule
\multirow{3}{*}{Llama 3.2 1B} & \textbf{7} & \textbf{0.2221} & \textbf{0.1637} & \textbf{0.1246} \\
& 11 & 0.2178 & 0.1595 & 0.1244 \\
& 15 & 0.2098 & 0.1589 & 0.1175 \\
\midrule
\multirow{5}{*}{Llama 3.2 3B} & 7 & 0.2217 & 0.1603 & 0.1231 \\
& \textbf{11} & \textbf{0.2274} & 0.1722 & \textbf{0.1303} \\
& 15 & 0.2251 & \textbf{0.1746} & 0.1219 \\
& 19 & 0.2264 & 0.1714 & 0.1195 \\
& 23 & 0.2196 & 0.1664 & 0.1141 \\
\midrule
\multirow{4}{*}{Whisper Large V3} & \textbf{12} & \textbf{0.1759} & 0.1073 & 0.0902 \\
& 25 & 0.1728 & 0.1111 & 0.0882 \\
& 31 & 0.1626 & 0.1056 & 0.0895 \\
& layernorm & 0.1739 & \textbf{0.1117} & \textbf{0.0937} \\
\bottomrule
\end{tabular}
\end{table}

\section{Context Length for Multimodal LLMs}
\label{sec:context_len_appd}
This study was conducted to determine the optimal temporal context window for feature extraction from the M-LLMs. The primary context length used in our final models was 20 seconds, a decision initially guided by the computational and time constraints of the challenge. To validate this choice and explore potential improvements, we extracted features using varied context lengths. For each model, these features were extracted exclusively from the single, most performant layer identified in our layer-wise ablation study.

For this specific ablation, we trained separate encoding models on a reduced corpus consisting of \textit{Friends S1} and \textit{Figures}. The performance of these models was then evaluated on a held-out in-distribution dataset (\textit{Friends S6}) and an out-of-distribution dataset (\textit{Life}). The results indicated that the 20-second window provided the most robust and highest overall performance across both M-LLM families. However, we noted a model-specific effect where the Qwen2.5-Omni model demonstrated a significant performance increase with a longer, 40-second context window. 

\begin{table}[h!]
\centering
\caption{M-LLM Performance Across Varied Context Lengths}
\label{tab:context_length_ablation}
\begin{tabular}{lccc}
\toprule
\textbf{Model} & \makecell[t]{\textbf{Context Length} \\ \textmd{(in seconds)}} & \multicolumn{2}{c}{\textbf{Pearson Correlation Score}} \\[-10pt] 
\cmidrule(lr){3-4}
& & \textit{Friends S6} & \textit{Life} \\
\midrule
\multirow{4}{*}{InternVL3-8B} & 10  & 0.2216 & 0.0948 \\
& \textbf{20} & \textbf{0.2331} & \textbf{0.1392} \\
& 30  & 0.2042 & 0.0899 \\
& 40  & 0.2052 & 0.0894 \\
\midrule
\multirow{4}{*}{InternVL3-14B} & 10 & \textbf{0.2161} & 0.0928 \\
& \textbf{20}  & 0.2082 & \textbf{0.0955} \\
& 30  & 0.2078 & 0.0939 \\
& 40  & 0.2081 & 0.0949 \\
\midrule
\multirow{4}{*}{Qwen2.5-Omni 3B} & 10  & 0.2432 & 0.1401 \\
& 20  & 0.2430 & 0.1433 \\
& 30 & 0.2466 & 0.1477 \\
& \textbf{40} & \textbf{0.2470} & \textbf{0.1517} \\
\midrule
\multirow{4}{*}{Qwen2.5-Omni 7B} & 10  & 0.2458 & \textbf{0.1541} \\
& 20  & 0.2315 & 0.1499 \\
& 30 & 0.2523 & 0.1521 \\
& \textbf{40}  & \textbf{0.2525} & 0.1512 \\
\bottomrule
\end{tabular}
\end{table}

\section{Effect of Validation Set on Ensemble Performance}
\label{appendix:validation_set}

The choice of validation set plays a crucial role in determining which models are selected for the ensemble and can significantly affect generalization to out-of-distribution (OOD) data. To illustrate this, we compared two ensembling strategies: one using only \textit{life} as the validation dataset, and another averaging performance across both \textit{life} and \textit{bourne}.

\begin{figure}[]
  \centering
  \includegraphics[width=0.7\textwidth]{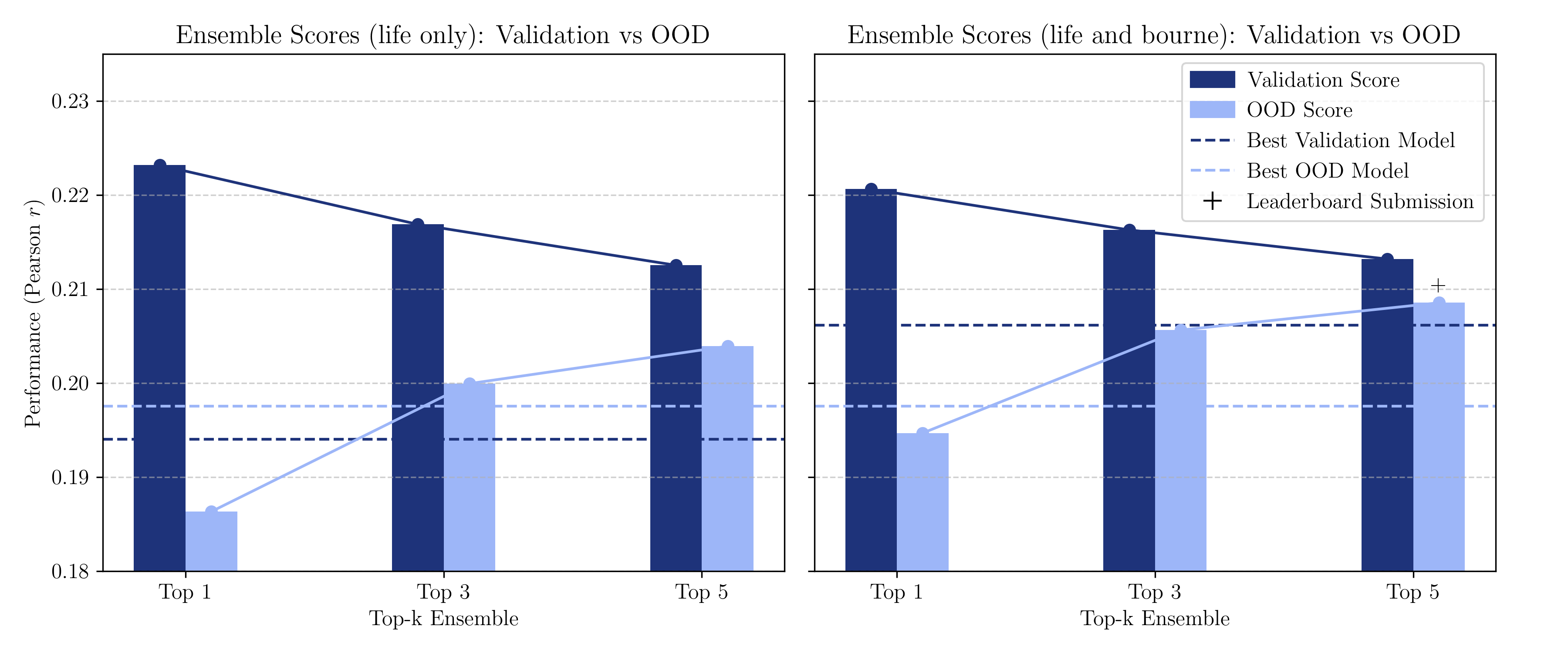}
  \caption{
    Impact of validation set selection on ensemble performance across validation and out-of-distribution (OOD) data. A total of 49 models were considered. The left panel shows ensembles selected using only \textit{life}, while the right panel uses the average score across \textit{life} and \textit{bourne}. Bars indicate Top-$k$ ensemble scores. Dashed lines represent the performance of the best single model, and the `+' symbol denotes the score submitted to the leaderboard.
  }
  \label{fig:ensemble_scores_life}
\end{figure}

As shown in Figure~\ref{fig:ensemble_scores_life}, both strategies yield similar expected (validation) performance, but the impact on OOD generalization is notable. When using only \textit{life} for validation, the Top-5 ensemble reached an OOD Pearson correlation of 0.2039. In contrast, incorporating both \textit{life} and \textit{bourne} increased the Top-5 OOD score to 0.2085. The gap is even more pronounced in the Top-3 and Top-1 ensembles, suggesting that more robust validation yields more transferable model selections.

These findings highlight the importance of choosing diverse and representative validation data when constructing ensembles—particularly in the presence of strong domain shifts between training and test distributions.

\section{Cross-subject encoding ceiling}
\label{sec:cross_subject}

\begin{figure}[]
  \centering
  \includegraphics[width=4.5in]{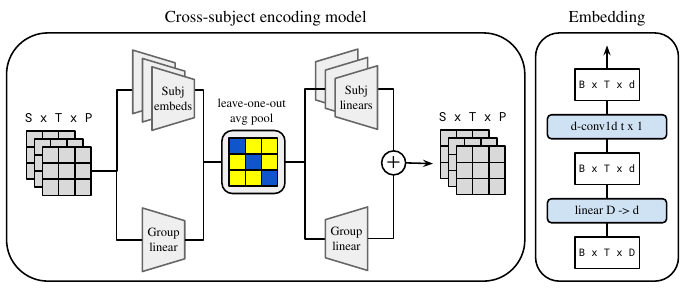}
  \caption{
    Cross-subject encoding model architecture. (\texttt{T} = number of time points, \texttt{S} = number of subjects, \texttt{P} = number of parcels (1000), \texttt{d} = embedding dimension, \texttt{B} = batch size.)
  }
  \label{fig:arch_cross}
\end{figure}

To estimate a performance ceiling for our feature encoding model, we compare it with a \textit{cross-subject} encoding model (Figure~\ref{fig:arch_cross}). The goal of the cross-subject encoding model is to jointly predict each subject's fMRI activity, using the activity from the remaining three other subjects as input.
Similar to the feature encoding model, the cross-subject model is made up of an embedding stage and a prediction stage. In the embedding stage, the input fMRI activity time series of each subject is projected to a small latent dimension using a shared group plus subject-specific residual linear projection, similar to the fMRI prediction head above. To filter and temporally align each subject's embedding time series, we apply subject-specific depth-wise 1D convolutions with a small kernel width. We next compute a pooled latent embedding for each subject by leave-one-subject-out average pooling, so that the latent embedding for subject $i$ is computed by averaging the embedded activity for the other subjects $j \neq i$. Finally, we reconstruct the input fMRI activity for each subject using the same prediction head architecture as described in Section~\ref{sec:arch}.

\begin{figure}[]
\centering
\includegraphics[height=2.5in]{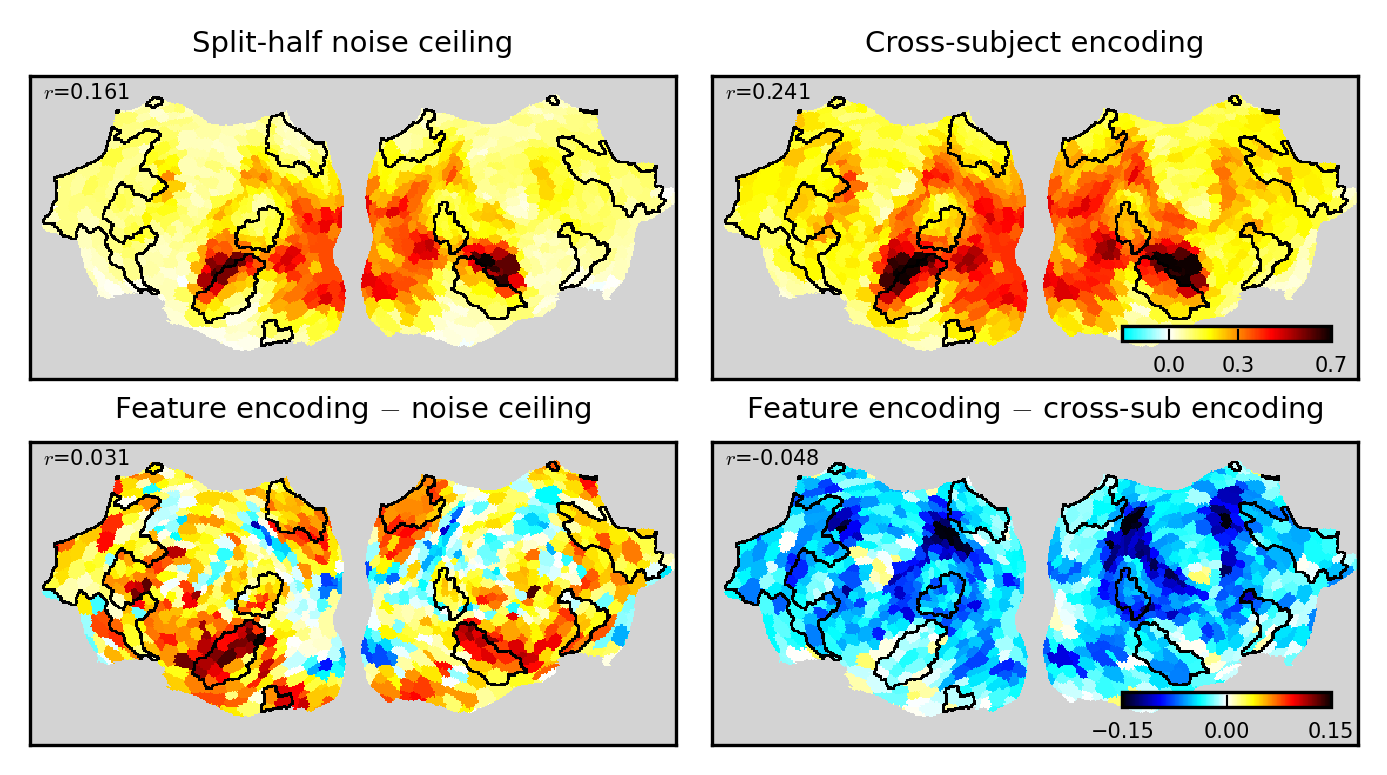}
\captionof{figure}{
Comparison of performance ceiling estimates on the movie \textit{life}.
Top row shows the Pearson accuracy maps for each method. The bottom row shows the difference between feature encoding accuracy and the ceiling.
}
\label{fig:noise_ceil}
\end{figure}

\myparagraph{Ceiling comparison.} In Figure~\ref{fig:noise_ceil} we compare our baseline model performance with two approaches for estimating a performance ``ceiling' on the movie \textit{life}. The first approach uses a simple split-half correlation estimate of the noise ceiling \citep{schrimpf2018brain}, taking advantage of the fact that a subset of the movies (\textit{figures}, \textit{life})  were shown twice. The second ceiling estimation uses our cross-subject encoding model described in Section~\ref{sec:arch}. The intuition is that the aligned fMRI activity from other subjects should be an ideal ``feature'' for predicting a target subject's activity.

We observe that our baseline model outperforms the simple split-half correlation noise ceiling in most brain areas\footnote{This is not so surprising. The split-half correlation is a conservative lower bound of the true noise ceiling, since the repeat measurement is a noisy estimate of the true expected stimulus driven activity. Some corrections have been suggested, e.g.~\citet{schoppe2016measuring}. However these provide at best a loose upper bound.}, except for parts of the peripheral early visual cortex and the dorsal attention network. By contrast, the cross-subject encoding model appears to be a more reliable ceiling estimate, outperforming the feature encoding model across the brain. In the difference map between the feature encoding and cross-subject encoding models, we observe significant unexplained variance in multiple brain areas, but especially in the dorsal attention network. A direction for future work is to investigate closing the gaps between feature encoding performance and the cross-subject encoding ceiling.

\end{document}